\documentclass[10pt,twocolumn,letterpaper]{article}

\usepackage{iccv}
\usepackage{times}
\usepackage{epsfig}
\usepackage{graphicx}
\usepackage{amsmath}
\usepackage{amssymb}
\usepackage{dsfont}
\usepackage{subfigure}
\usepackage{multirow}

\usepackage[breaklinks=true,bookmarks=false]{hyperref}

\iccvfinalcopy 

\def\iccvPaperID{****} 

\begin{document}

\title{Towards Effective Codebookless Model for Image Classification}

\author{Qilong Wang$^{1}$, Peihua Li$^{1}$, Lei Zhang$^{2}$, Wangmeng Zuo$^{3}$\\
$^{1}$Dalian University of Technology, $^{2}$Hong Kong Polytechnic University, $^{3}$Harbin Institute of Technology\\
{\tt\small qlwang@mail.dlut.edu.cn, peihuali@dlut.edu.cn, cslzhang@comp.polyu.edu.hk, wmzuo@hit.edu.cn}
}

\maketitle

\begin{abstract}
   The bag-of-features (BoF) model for image classification has been thoroughly studied over the last decade. Different from the widely used BoF methods which modeled images with a pre-trained codebook, the alternative codebook-free image modeling method, which we call Codebookless Model (CLM), attracted little attention. In this paper, we present an effective CLM that represents an image with a single Gaussian for classification. By embedding Gaussian manifold into a vector space, we show that the simple incorporation of our CLM into a linear classifier achieves very competitive accuracy compared with state-of-the-art BoF methods (e.g., Fisher Vector). Since our CLM lies in a high-dimensional Riemannian manifold, we further propose a joint learning method of low-rank transformation with support vector machine (SVM) classifier on the Gaussian manifold, in order to reduce computational and storage cost. To study and alleviate the side effect of background clutter on our CLM, we also present a simple yet effective partial background removal method based on saliency detection. Experiments are extensively conducted on eight widely used databases to demonstrate the effectiveness and efficiency of our CLM method.
\end{abstract}

\section{Introduction}

\begin{figure}[htb]
\begin{center}
   \includegraphics[width=1.0\linewidth]{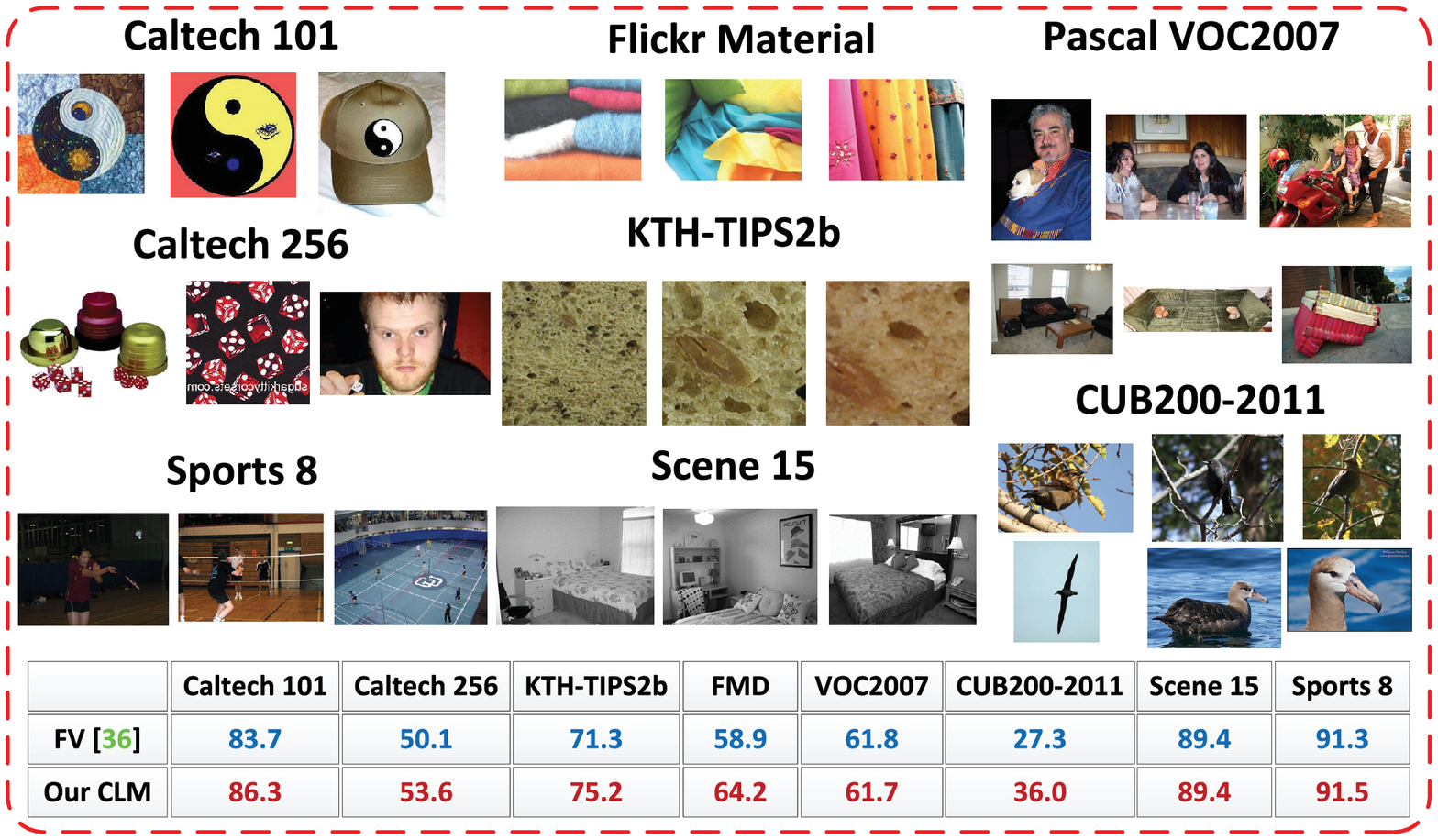}
\end{center}
\caption{Some example images and comparison (in \%) between Fisher vector (FV) and our codebookless model (CLM) on various image databases.}
\label{fig:motivation}
\end{figure}

Image classification has been attracting massive attentions in computer vision and pattern recognition communities in recent years. It is one of the most fundamental but challenging vision problems because images, as illustrated in
Fig.~\ref{fig:motivation}, often suffer from significant scale, view or illumination variations (e.g., in texture classification \cite{CaputoHM05} and material recognition \cite{FMD}), and pose changes, background clutter, partial occlusion (e.g., in scene categorization \cite{Lazebnik:2006:BBF,LiF07} and object recognition \cite{Everingham10,TPAMI2006,Caltech256,WahCUB2002011}).

\begin{figure*}
\begin{center}
\includegraphics[width=0.9\linewidth]{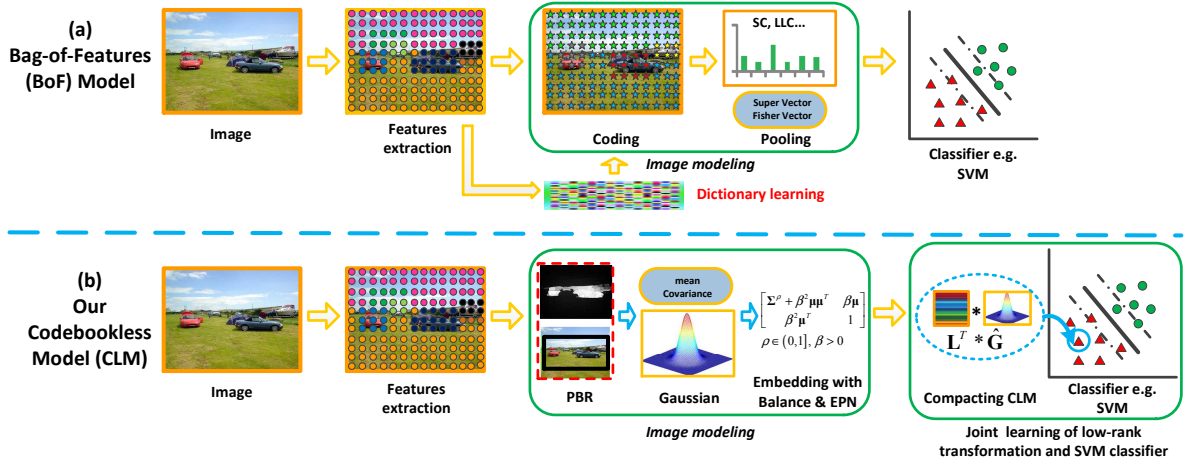}
\end{center}
   \caption{Comparison between (a) the BoF model and (b) our CLM. The major difference between them is that whether there is a pre-trained codebook \& coding or not. Our CLM mainly consists of a Gaussian model for image representation and a joint low-rank learning with linear SVM classifier.}
\label{fig:overview}
\end{figure*}

For a long time the bag-of-features (BoF) model \cite{Sivic03} has been almost given priority to image classification. As shown in Fig.~\ref{fig:overview} (a), the BoF-based methods generally consist of five components: local features extraction, learning codebook with training data, coding local features with pre-trained codebook, pooling or aggregating codes over images, and finally, learning classifier (e.g., SVM) for classification. With this processing pipeline, the BoF-based methods can be seen as a hand-crafted five-layer hierarchical feed-forward network \cite{Sydorov_2014_CVPR} with a pre-trained feature coding template (codebook) \cite{NIPS2014_5286}. The learned codebook depicts the distribution of feature space, and makes coding of high dimensional features possible. This architecture has achieved very promising performance in a variety of image classification tasks.

The codebook as a reference for feature coding serves as a bridge between local features and global image representation. However, it is well known that segmentation of feature space involved in building of codebook brings on quantization error \cite{Irani-Definese-CVPR2008}, and leads to continuous striving for this side effect (e.g., soft coding methods \cite{sanchez,GemertVSG10} alleviate but cannot completely eliminate it). Though offline, training of codebook,  particularly large size ones, is time consuming. In addition, in general the pre-trained codebook on one database cannot naturally adapt to other databases \cite{ZhouYLWLT14}.

An alternative approach is to estimate the statistics directly on sets of local features from input images \cite{carreira_pami14,Nakayama-CVPR2010,tuzel:region}, as illustrated in Fig.~\ref{fig:overview} (b),
which is called codebookless model (CLM) in this paper. It is clear from Fig.~\ref{fig:overview} that the major difference is that the BoF model learns a codebook to explore the statistical distribution of local features and then performs coding of descriptors, while the CLM represents images with descriptors directly, requiring no pre-trained codebook and the subsequent coding. Conceptually, the codebookless model has the potential to circumvent the aforementioned limitations of the BoF model, however, which has received little attention in image classification community. The main reasons may be that such methods have not yet shown competitive classification performance, and that they often need to utilize inefficient and unscalable kernel-based classifiers.

In this paper, we propose an effective CLM scheme, and argue that the CLM can be a competitive alternative to the BoF methods for image classification. The comparison between state-of-the-art BoF method, Fisher Vector (FV) \cite{sanchez}, and our CLM on various image databases is shown in Fig. \ref{fig:motivation}. First and foremost, we extract a set of local features (e.g., SIFT \cite{Lowe04}) on a dense grid of image, and simply model them with a single Gaussian model to represent the input image. Then, we employ a two-step metric for matching Gaussian models. By using this metric, Gaussian models can be fed to a linear classifier for ensuring efficient and scalable classification while respecting the Riemannian geometry structure of Gaussian models. Moreover, we introduce two well-motivated parameters into the used metric. One is to balance the effect between mean and covariance of Gaussian, and another is for eigenvalue power normalization on covariance.

Our codebookless model usually is of high dimension, by incorporating low-rank learning with SVM, we propose a joint learning method to effectively compress Gaussian models while respecting their Riemannian geometry structure. It is mentionable that, to the best of our knowledge, we make the first attempt to perform joint learning of low-rank transformation and SVM on Gaussian manifold. Finally, to alleviate the side effect of background clutter, a saliency-based partial background removal method is proposed to enhance our CLM. The experimental results show that partial background removal is helpful to CLM when images are heavily cluttered (e.g., CUB200-2011 and Pascal VOC2007).



\section{Related work}\label{related}

The codebookless model for directly modeling the statistics of local features has been studied in past decades. Rubner \etal \cite{rubner} introduced signatures for image representation, and proposed the Earth Mover's Distance for image matching which is robust but has high computational cost. Tuzel \etal \cite{tuzel:region} for the first time used covariance matrices for representing regular image regions, and employed Affine-Riemannian metric which suffers from high computational cost \cite{LogMetricsIJCV06}. Gaussian model as image descriptor has been used for visual tracking \cite{GongWang2009}, in which Gaussian models are matched based on the Riemannian metric, involving expensive operations to solve generalized eigenvalue problem. Going beyond Gaussian, Gaussian mixture model (GMM) is more informative and is used in image retrieval \cite{BIK11}. However, GMM suffers from some limitations, such as high computational cost of matching methods and lacking of general criteria for model selection.

Our work is motivated by \cite{carreira_eccv12,carreira_pami14} and \cite{Nakayama-CVPR2010}. Carreira \etal \cite{carreira_eccv12,carreira_pami14} modeled the free-form regions obtained by image segmentation with estimating  the second-order moments. By using Log-Euclidean metric \cite{Arsigny:2005}, the method in \cite{carreira_eccv12,carreira_pami14} can be combined with a linear classifier, which has shown competing recognition performance on images with less background clutter (e.g., Caltech101 \cite{TPAMI2006}). Different from \cite{carreira_eccv12,carreira_pami14}, we employ a Gaussian model to represent the whole image. It is well-known that a covariance matrix can be seen as a Gaussian model with fixed mean vector. Compared to \cite{carreira_eccv12,carreira_pami14}, our CLM contains both the first-order (mean) and second-order (covariance) information. Note that the first-order statistics has proven important in image classification \cite{JDSP10,sanchez}. Moreover, the manifold of Gaussian models and that of covariance matrices are quite different, and the embedding method in our CLM makes Gaussian models can be handled flexibly and conveniently.

Nakayama \etal \cite{Nakayama-CVPR2010} also represented an image with a global Gaussian for scene categorization. However, they matched two Gaussian models by using the Kullback-Leibler (KL) divergence, and hence kernel-based classifiers have to be used. This method is not scalable and has high computational cost. In contrast to \cite{Nakayama-CVPR2010}, our metric is decoupled which allows a linear classifier to be combined, which makes our method more efficient and scalable than the KL kernel based one in \cite{Nakayama-CVPR2010}. Moreover, compared with the ad-hoc linear kernel (Euclidean baseline) in \cite{Nakayama-CVPR2010}, our method takes advantage of the geometry structure of Gaussian models and brings large performance improvement.

There is another line of research on codebookless model methods. Grauman \etal \cite{Grauman05thepyramid} proposed a pyramid match kernel to map feature sets to multi-resolution histograms, and employed histogram intersection kernel for classification. Bo \etal \cite{NIPS2009_3874} presented efficient match kernels to map local features into a low dimensional space, and adopted a linear classifier. Boiman \etal \cite{Irani-Definese-CVPR2008} developed an image-to-class distance between the sets of local features, and employed a nearest neighbor classifier. Yao \etal \cite{YaoEtalCVPR12} proposed a codebook-free approach by using a large number of randomly generated image templates for image representation, and developed a bagging-based classifier.

\section{Proposed method}\label{AIMDL}
We first introduce the image representation by a single Gaussian model. Then, we employ an effective and efficient two-step metric for matching Gaussian models, and propose two well-motivated parameters to improve the used distance metric. Finally, we present a joint learning method of low-rank transformation and SVM on Gaussian manifold.

\subsection{Gaussian model for image representation}
Given an input image, we extract a set of $N$ local features $\{\mathbf{x}_{i} \in\mathbb{R}^{k\times1}, i=1,\ldots, N\}$ at a dense grid.
By the maximum likelihood method, the image can be  represented by the following Gaussian model:
\begin{align*}
\mathcal{N}(\mathbf{x}_{i}|\boldsymbol{\mu},\boldsymbol{\Sigma}) = \frac{\exp \big(-\frac{1}{2}(\mathbf{x}_{i}-\boldsymbol{\mu})^{T}\boldsymbol{\Sigma}^{-1}(\mathbf{x}_{i}-\boldsymbol{\mu}) \big)}{\sqrt{(2\pi)^{k}\mathrm{det}(\mathbf{\Sigma})}},\nonumber
\end{align*}
where $\boldsymbol{\mu}=\frac{1}{N}\sum_{i=1}^{N}\mathbf{x}_{i}$ and  $\boldsymbol{\Sigma}=\frac{1}{N-1}\sum_{i=1}^{N}(\mathbf{x}_{i}-\boldsymbol{\mu})(\mathbf{x}_{i}-\boldsymbol{\mu})^{T}$ are mean vector and covariance matrix, and $\mathrm{det}(\cdot)$ denotes matrix determinant. Compared with histogram
and covariance, Gaussian model is more informative. Meanwhile, unlike matching of signatures
\cite{rubner} or GMMs \cite{BIK11}, matching of Gaussian models does not bring high computational cost.

\subsection{Two-step metric between Gaussian models}

To match Gaussian models, we exploit a two-step metric which has been proposed to compute the ground distance between Gaussian components of GMMs \cite{LiWZ13}. The first step is to embed Gaussian manifold into the space of SPD matrices \cite{RePEcjmvana}, and then map the Lie group of SPD matrices into its corresponding Lie algebra, a linear space, by using the Log-Euclidean metric \cite{Arsigny:2005}.

The space of $k$-dimensional Gaussian models is a Riemannian manifold. Let $\mathcal{N}(\boldsymbol{\mu},\boldsymbol{\Sigma})$ be a Gaussian model with mean vector $\boldsymbol{\mu}$ and covariance matrix $\boldsymbol{\Sigma}$. Through a continuous function $\pi$,  $\mathcal{N}(\boldsymbol{\mu},\boldsymbol{\Sigma})$ is mapped to an affine matrix, an element in the affine group $\mathcal{A}_{k}^{+}=\{(\boldsymbol{\mu},\mathbf{P})|\boldsymbol{\mu} \in\mathbb{R}^{k\times1}, \mathbf{P}\in\mathbb{R}^{k\times k},\det(\mathbf{P})>0\}$; that is,
\begin{align}\label{equ:embeding1}
\pi: \mathcal{N}(\boldsymbol{\mu},\boldsymbol{\Sigma}) \mapsto \mathbf{A} = \begin{bmatrix}
\mathbf{P} & \boldsymbol{\mu} \\
\mathbf{0}^{T} & 1\\
\end{bmatrix},
\end{align}
where $\boldsymbol{\Sigma}=\mathbf{P}\mathbf{P}^{T}$ is the Cholesky factorization of $\boldsymbol{\Sigma}$. Further, through the function $\gamma: \mathbf{A}\mapsto \mathbf{S}=\mathbf{A}\mathbf{A}^{T}$, $\mathbf{A}$ is mapped to an SPD matrix $\mathbf{S}$. So far, by the successive functions $\pi$ and $\gamma$, $\mathcal{N}(\boldsymbol{\mu},\boldsymbol{\Sigma})$ is uniquely designated as an $(k+1)\times(k+1)$ SPD matrix
\begin{align}\label{equ:embeding2}
\mathcal{N}(\boldsymbol{\mu},\boldsymbol{\Sigma}) \sim \mathbf{S}= \begin{bmatrix}
\boldsymbol{\Sigma}+ \boldsymbol{\mu}\boldsymbol{\mu}^{T}& \boldsymbol{\mu} \\
\boldsymbol{\mu}^{T} & 1\\
\end{bmatrix}.
\end{align}
Please refer to \cite{RePEcjmvana} for details on the embedding process.

The space of $(k+1)\times(k+1)$ SPD matrices $\mathcal{S}_{k+1}^{+}$ is a Lie group that forms a Riemannian manifold.
Two operations, namely the logarithmic multiplication and the scalar logarithmic multiplication, are defined in the Log-Euclidean metric \cite{Arsigny:2005}, which equip $\mathcal{S}_{k+1}^{+}$ with structures of not only the Lie group but also vector space. Through the matrix logarithm, $\mathcal{S}_{k+1}^{+}$ is mapped into its Lie algebra $\mathcal{S}_{k+1}$, the vector space of $(k+1)\times(k+1)$ symmetric matrices. The matrix logarithm is a deffemorphism and an isomorphism so that operations over SPD matrices can be replaced by the Euclidean operations of their counterparts in the vector space. So, through the matrix logarithm, an SPD matrix $\mathbf{S}$ is one-to-one mapped to a symmetric matrices $\mathbf{G}$  which lies in a linear space, and the geodesic distance between SPD matrices $\mathbf{S}_{i}$ and $\mathbf{S}_{j}$ is defined by $dist_{\mathbf{S}_{i}, \mathbf{S}_{j}}=\|\mathbf{G}_{i}-\mathbf{G}_{j}\|_{F}$, where $F$ is the Frobenius norm.

\subsection{Two well-motivated parameters}

In practice, we found that it is important to balance mean vector and covariance matrix in the embedding matrix (\ref{equ:embeding2}), because their dimensions and order of magnitude of each dimension may vary considerably. Meanwhile, the effect of mean vector and covariance matrix may vary for different tasks. With these considerations, we introduce a parameter $\beta>0$ in the function $\pi$ (\ref{equ:embeding1}):
\begin{align}\label{equ:embeding3}
\pi(\beta):\;\mathcal{N}(\boldsymbol{\mu},\boldsymbol{\Sigma}) \mapsto \mathbf{A} = \begin{bmatrix}
\mathbf{P} & \beta \boldsymbol{\mu} \\
\mathbf{0}^{T} & 1\\
\end{bmatrix}.
\end{align}
Accordingly, the embedding matrix has the following form:
\begin{align}\label{equ:embeding4}
\mathcal{N}(\boldsymbol{\mu},\boldsymbol{\Sigma}) \sim \mathbf{S}(\beta)= \begin{bmatrix}
\boldsymbol{\Sigma}+ \beta^{2}\boldsymbol{\mu}\boldsymbol{\mu}^{T}& \beta \boldsymbol{\mu} \\
\beta \boldsymbol{\mu}^{T} & 1\\
\end{bmatrix}.
\end{align}
The embedding matrix (\ref{equ:embeding4}) reduces to the covariance matrix when $\beta=0$, and is equal to the original one when $\beta=1$. Hence, the role of mean vector and covariance matrix can be adjusted by $\beta$.

The maximum likelihood estimator of the empirical covariance matrix is susceptible to interference of noise, especially for high dimension space \cite{Donoho2014}. Based on observation that the maximum likelihood estimator of covariance ought to be improvable by eigenvalue shrinkage \cite{Stein1986}, we exploit power normalization on the eigenvalues of covariance matrix (EPN). Let $\mathcal{N}(\boldsymbol{\mu},\boldsymbol{\Sigma})$ be a Gaussian model estimated from a set of descriptors extracted from some image. The covariance matrix $\boldsymbol{\Sigma}$ has eigenvalue decomposition $\boldsymbol{\Sigma} = \mathbf{U} \mathrm{diag} (\lambda_{i})\mathbf{U}^{T}$, where $\mathbf{U}$ is an orthornormal matrix whose $i^{\mathrm{th}}$ column is the eigenvector of $\boldsymbol{\Sigma}$ and $\lambda_{i}>0$ is the corresponding eigenvalue, and $\mathrm{diag}(\cdot)$ denotes diagonal matrix. Then by introducing a parameter $\rho$, our normalization is defined as
\begin{align}\label{EPN}
\boldsymbol{\Sigma}^{\rho}  = \mathbf{U}\mathrm{diag}(\lambda_{i}^{\rho})\mathbf{U}^{T},\; \text{with} \;0<\rho\leq 1.
\end{align}
With EPN, our final embedding matrix is:
\begin{align}\label{equ:embeding5}
\mathcal{N}(\boldsymbol{\mu},\boldsymbol{\Sigma}) \sim \mathbf{S}(\beta,\rho)= \begin{bmatrix}
\boldsymbol{\Sigma}^{\rho}+ \beta^{2}\boldsymbol{\mu}\boldsymbol{\mu}^{T}& \beta \boldsymbol{\mu} \\
\beta \boldsymbol{\mu}^{T} & 1\\
\end{bmatrix}.
\end{align}
It is easy to prove that the embedding matrix (\ref{equ:embeding5}) is still positive definite as $\boldsymbol{\Sigma}^{\rho}$ being an SPD matrix. The eigenvalues power normalization has been proposed to measure distances between covariance matrices \cite{Dryden2009,Jayasumana_2013} or tensor \cite{koniusz:hal-00922524}, namely, Power-Euclidean metric. Different from previous work, we use eigenvalues power normalization for robust estimation of covariance matrices in Gaussian setting for the case of high dimensional features, and compare Gaussians by using Gaussian embedding and the Log-Euclidean metric.

According to the Log-Euclidean framework, the matrix $\mathbf{S}(\beta, \rho)$ can be further embedded into a linear space by matrix logarithm:
\begin{equation}\label{equ:embed-linear}
\mathbf{G}(\beta,\rho)=\log(\mathbf{S}(\beta, \rho)).
\end{equation}
Let $\mathcal{N}_{i}=\mathcal{N}(\boldsymbol{\mu}_{i},\boldsymbol{\Sigma}_{i})$ and $\mathcal{N}_{j}=\mathcal{N}(\boldsymbol{\mu}_{j},\boldsymbol{\Sigma}_{j})$ be
two Gaussian models and their corresponding symmetric matrices are $\mathbf{G}_{i}(\beta,\rho)$ and $\mathbf{G}_j(\beta,\rho)$.
The distance between two Gaussian models is
\begin{equation}\label{distance_M_L}
dist_{\mathcal{N}_{i},\mathcal{N}_{j}}=\big\|\mathbf{G}_{i}(\beta,\rho)-\mathbf{G}_{j}(\beta,\rho)\big\|_{F}.
\end{equation}
It is easy to know that distance (\ref{distance_M_L}) is decoupled so that $\mathbf{G}_{i}(\beta,\rho)$ and $\mathbf{G}_{j}(\beta,\rho)$ can be computed separately and adopted in a linear classifier. For notational simplicity, we omit the parameters $\beta$ and $\rho$ in the distance measure (\ref{distance_M_L}).

\subsection{Joint low-rank learning and SVM classifier}
Our CLM usually is of high dimension ($>10^{4}$). In order to suppress redundant and noisy information while reducing computational and storage cost, we propose a low-rank learning method to compact our CLM. The matrix $\mathbf{G}$ in geodesic distance (\ref{distance_M_L}) is a $(k+1)\times(k+1)$ symmetric matrix which lies in the Euclidean space. Due to its symmetry, we can unfold the upper triangular part of $\mathbf{G}$ to a vector of size  $d=(k+1)\times(k+2)/2$. We can modify geodesic distance (\ref{distance_M_L}) by introducing a low-rank transformation matrix $\mathbf{L}\in\mathbb{R}^{d\times r}, r\ll d$:
\begin{align}
dist_{\mathcal{N}_{i},\mathcal{N}_{j}}=\|\mathbf{L}^{T}(\mathbf{f}_{i}-\mathbf{f}_{j})\|_{2},
\end{align}
where $\mathbf{f}_{i}$ and $\mathbf{f}_{j}$ are the unfolding vectors of two Gaussian models $\mathcal{N}_{i}$ and $\mathcal{N}_{j}$, respectively.


Recent researches \cite{JiY09,Weston2011} have shown that joint optimization of dimensionality reduction with classifier performs better than separate optimization of the two modules. Thus, given $N$ training samples $\{\mathbf{f}_{n},n\in [1,N]\}$, we optimize the low-rank learning jointly with a linear SVM (LRSVM):
\begin{align}\label{LRSVM2}
& \min\limits_{\mathbf{L},\mathbf{w},\boldsymbol{\xi}} \frac{1}{2} \|\mathbf{w}\|^{2} + C \sum_{n=1}^{N} \xi_{n} \\
& s.t. \quad  y_{n}(\mathbf{w}^{T}\mathbf{L}^{T}\mathbf{f}_{n}+b)\geq 1-\xi_{n}, \forall \xi_{n}>0, n\in [1,N], \nonumber \\
&\quad \quad  \mathbf{L}^{T}\mathbf{L} = \mathbf{I}, \nonumber
\end{align}
where $\mathbf{w},\boldsymbol{\xi},b$ are parameters of SVM, and $y_{n}$ is the label of $\mathbf{f}_{n}$. The dimensionality reduction for SPD matrices \cite{Mehrtash14} has been studied with dimensionality reduction and classification separately performed, while our method is quite different in that we focus on Gaussian models and perform joint learning of low-rank transformation and SVM.

In practice, we extend the objective function (\ref{LRSVM2}) to multi-class problem under the spatial pyramid matching (SPM) framework \cite{Lazebnik:2006:BBF}. Given an image $I_{n}$, we can obtain its SPM representation $\mathbf{F}_{n} = [(\mathbf{f}_{n}^{1})^{T},\ldots,(\mathbf{f}_{n}^{B})^{T}]^{T}$ , where $B$ is the number of blocks in SPM, which is fed to a one vs. all SVM for solving the $M$ classes problem. As suggested in \cite{JiY09}, we optimize the dual problem of the objective function (\ref{LRSVM2}) under the SPM framework:
\begin{align}\label{LRSVM4}
& \min\limits_{\widehat{\mathbf{L}}}\max\limits_{\boldsymbol{\alpha}_{m}} \sum_{m=1}^{M} \big( \sum_{n=1}^{N} \alpha_{m}^{n}-\frac{1}{2} ( \boldsymbol{\alpha}_{m}^{T}\mathbf{Y}_{m}\mathbf{F}\mathbf{H}\mathbf{F}^{T}
\mathbf{Y}_{m}\boldsymbol{\alpha}_{m})\big) \nonumber \\
& s.t. \quad  \sum_{n=1}^{N} y_{m}^{n}\alpha_{m}^{n}=0, 0\leq \boldsymbol{\alpha}_{m} \leq C, \forall m \\
&\quad \quad  \widehat{\mathbf{L}}^{T}\widehat{\mathbf{L}} = \mathbf{I}, \; \widehat{\mathbf{L}}^{T}= \mathrm{Diag}(\mathbf{L}_{1}^{T},\ldots,\mathbf{L}_{B}^{T}),\; \mathbf{H} = \widehat{\mathbf{L}}\widehat{\mathbf{L}}^{T}, \nonumber
\end{align}
where $\mathbf{F} = [\mathbf{F}_{1},\ldots,\mathbf{F}_{N}]^{T}$ indicates all training features, and $\mathbf{Y}_{m}$ is the diagonal label matrix of the $m$th class with diagonal element $\mathbf{Y}_{m}(n,n)=y_{m}^{n}$.

The problem (\ref{LRSVM4}) is non-convex and can be optimized by a two-step alternating method: \emph{Step One}, fixing $\widehat{\mathbf{L}}$, we can optimize the Lagrange parameters $\boldsymbol{\alpha}_{m}$ with off-the-shelf SVM; \emph{Step Two}, for fixed $\boldsymbol{\alpha}_{m}$, we solve the following trace maximization problem:
\begin{align}\label{LRSVM5}
&\max\limits_{\widehat{\mathbf{L}}} \mathrm{tr}\bigg(\widehat{\mathbf{L}}^{T}\mathbf{F}^{T}\sum_{m=1}^{M}  (\mathbf{Y}_{m}\boldsymbol{\alpha}_{m} \boldsymbol{\alpha}_{m}^{T} \mathbf{Y}_{m}^{T})\mathbf{F}\widehat{\mathbf{L}}\bigg) \\ \nonumber
&s.t. \quad  \widehat{\mathbf{L}}^{T}\widehat{\mathbf{L}} = \mathbf{I}, \; \widehat{\mathbf{L}}^{T}= \mathrm{Diag}(\mathbf{L}_{1}^{T},\ldots,\mathbf{L}_{B}^{T}). \nonumber
\end{align}
We optimize the problem (\ref{LRSVM5}) by independently solving each $\mathbf{L}_{i}^{T}, \; i=1,\ldots,B$ with a close-form solution \cite{JiY09}. Due to the problem (\ref{LRSVM4}) being non-convex, initialization is nontrivial to reach a good local optimal solution and for fast convergence. In this paper, we use the basis of principal component analysis (PCA) as initialization, and we find that it can always achieve good performance and fast convergence.


\section{Partial background removal (PBR)}

We then present a simple yet effective method for analyzing and handling the side effect of background clutter based on unsupervised, bottom-to-up saliency detection. Our purpose here is to remove the interference of background, which is distinguished from the purpose of precise foreground localization in saliency detection community. Our method consists of two steps: coarse foreground detection and partial background removal. In the first step we localize in image the foreground based on saliency detection method \cite{Jiang_2013_ICCV} and then determine the bounding-box surrounding the foreground. Next, we adaptively expand bounding-box to accommodate some background regions based on size and intensity variance of the area inside the bounding-box. Then, the area outside bounding-box is removed for recognition. Our method is based on the considerations that accurate foreground detection is currently very difficult and neighboring regions of object can serve as the context and may be helpful for recognition. In our experiments, we adopt PBR to the two datasets with heavy background clutter: CUB200-2011 and VOC2007. Since PBR is designed for foreground objects with separable background clutter, we do not perform PBR on images with less background clutter and scene images where both foreground and background are valuable for scene understanding.

\section{Implementation details}\label{ID}
We extract multi-scale SIFT descriptors \cite{Lowe04} (standard pipeline in the BoF model) with cell size $2^{i}$, $i=1,2,\ldots$, and single scale pixel-wise covariance descriptor \cite{tuzel:region} via the dense sampling strategy with step-length 2.  The dense covariance descriptors are computed with 17 dimensional raw features including intensity and four kinds of first-order and second-order gradients from \cite{Pratt2007}. We perform matrix logarithm on the covariance descriptors (LogCov), which are then vectorized. The SIFT features are calculated via the VLFeat library \cite{vedaldi08vlfeat}. Moreover, following \cite{carreira_eccv12,carreira_pami14}, we also extract additional image cues, including color, location, scale, gradient and entropy to concatenate SIFT and LogCov. In order to ensure that there is sufficient data to estimate Gaussian models and covariance matrices are positive definite, we limit the minimum size of width or height of images to be larger than 64, and add $10^{-3}$ to the diagonal entries of covariance matrices, respectively. We employ the spatial pyramid strategy \cite{Lazebnik:2006:BBF} which divides an image into some regular regions (e.g., $1\times1$, $2\times2$, $1\times3$, $4\times4$). For each region we compute a Gaussian model, and then concatenate them to represent the whole image. Each Gaussian is weighted by $\frac{1/N_{l}}{\sum_{l=1}^{L}1/N_{l}}$, where $L$ and $N_{l}$ are the number of pyramid levels and regions in the $l^{\mathrm{th}}$ layer, respectively. We implement a one-vs-all SVM with LibSVM \cite{Chang:2011} and set parameter $C$ to $0.01$ on VOC2007 and $10$ on all the other databases. All algorithms are written in Matlab, and run on a PC equipped with i7-4770k CPU and 32G RAM.

\section{Experimental evaluation}\label{EE}
In this section, we evaluate the classification performance of our CLM on eight benchmark databases. First of all,
we make an analysis of local features, the parameters of our method, the proposed low-rank learning method and the partial background removal method on the challenging CUB200-2011 \cite{WahCUB2002011}. Then, we compare with state-of-the-art methods on Caltech101 \cite{TPAMI2006}, Caltech256 \cite{Caltech256}, KTH-TIPS2b \cite{CaputoHM05}, Flickr Material Database (FMD) \cite{FMD}, Pascal VOC2007 \cite{Everingham10}, Scene15 \cite{Lazebnik:2006:BBF} and Sports8 \cite{LiF07}. Finally, we analyze the computational complexity of our CLM.
\subsection{Parameters analysis}\label{Pa}

\begin{table}[htb]
\scriptsize
\begin{center}
\begin{tabular}{|l|c|c|c|c|c|c|c|c|c|}
\hline
 & \multicolumn{4}{|c|}{Local descriptors}  & \multicolumn{2}{|c|}{Parameters} & \multicolumn{2}{|c|}{BR} & \\
\hline
 & ST  & eST & LC & eLC & Beta & EPN & PBR & GT & Acc. \\
\hline
\multirow{2}{*}{Cov.} & $\checkmark$ &  &  &  &  & & & &$16.8$\\
\cline{2-6}\cline{7-10}
&  & $\checkmark$ &  &  & & & & & $24.1$ \\
\hline
\multirow{11}{*}{Gau.} & $\checkmark$ &  &  &  &  & & & &$18.6$\\
\cline{2-6}\cline{7-10}
&  & $\checkmark$ &  &  & & & & & $25.6$ \\
\cline{2-6}\cline{7-10}
&  &  & $\checkmark$ &  & & & & & $19.1$ \\
\cline{2-6}\cline{7-10}
&  &  &  & $\checkmark$ & & & & & $26.3$ \\
\cline{2-6}\cline{7-10}
&  & $\checkmark$ &  &  & $\checkmark$ & & & & $26.5$ \\
\cline{2-6}\cline{7-10}
&  & $\checkmark$ &  &  &  & $\checkmark$ & & & $26.8$ \\
\cline{2-6}\cline{7-10}
&  & $\checkmark$ &  &  &  &  & $\checkmark$ & & $33.3$ \\
\cline{2-6}\cline{7-10}
&  & $\checkmark$ &  &  &  &  & & $\checkmark$ & $45.3$ \\
\cline{2-6}\cline{7-10}
&  & $\checkmark$ &  &  & $\checkmark$ & $\checkmark$ & & & $28.1$ \\
\cline{2-6}\cline{7-10}
&  & $\checkmark$ &  &  & $\checkmark$ & $\checkmark$ & $\checkmark$ & & $36.0$ \\
\cline{2-6}\cline{7-10}
&  & $\checkmark$ &  &  & $\checkmark$ & $\checkmark$ &  & $\checkmark$ & $48.2$ \\
\hline
\end{tabular}
\end{center}
\caption{Classification results (in \%) of our CLM vs. various combinations of descriptors, parameters and background removal on CUB200-2011.}
\label{table:parameters}
\end{table}

\noindent \textbf{Local descriptors}\; Four kinds of local descriptors, SIFT (ST) and its enrichment (eST), and LogCov (LC) and its enrichment (eLC), are evaluated in this section. The results of our CLM with various local descriptors on CUB200-2011 are shown in Table \ref{table:parameters}. We can see that the Gaussian model used in our method outperforms covariance matrix by $1.5\%$ or higher with either SIFT or eSIFT,  which, we believe, can indicate that the first-order (mean) information is non-trivial. We use eST to evaluate other parameters as follows.

\begin{figure}[htb]
\centering
\subfigure{\label{figure:balance} 
\begin{minipage}[b]{0.5\linewidth}
\centering
\includegraphics[width=1.0\textwidth]{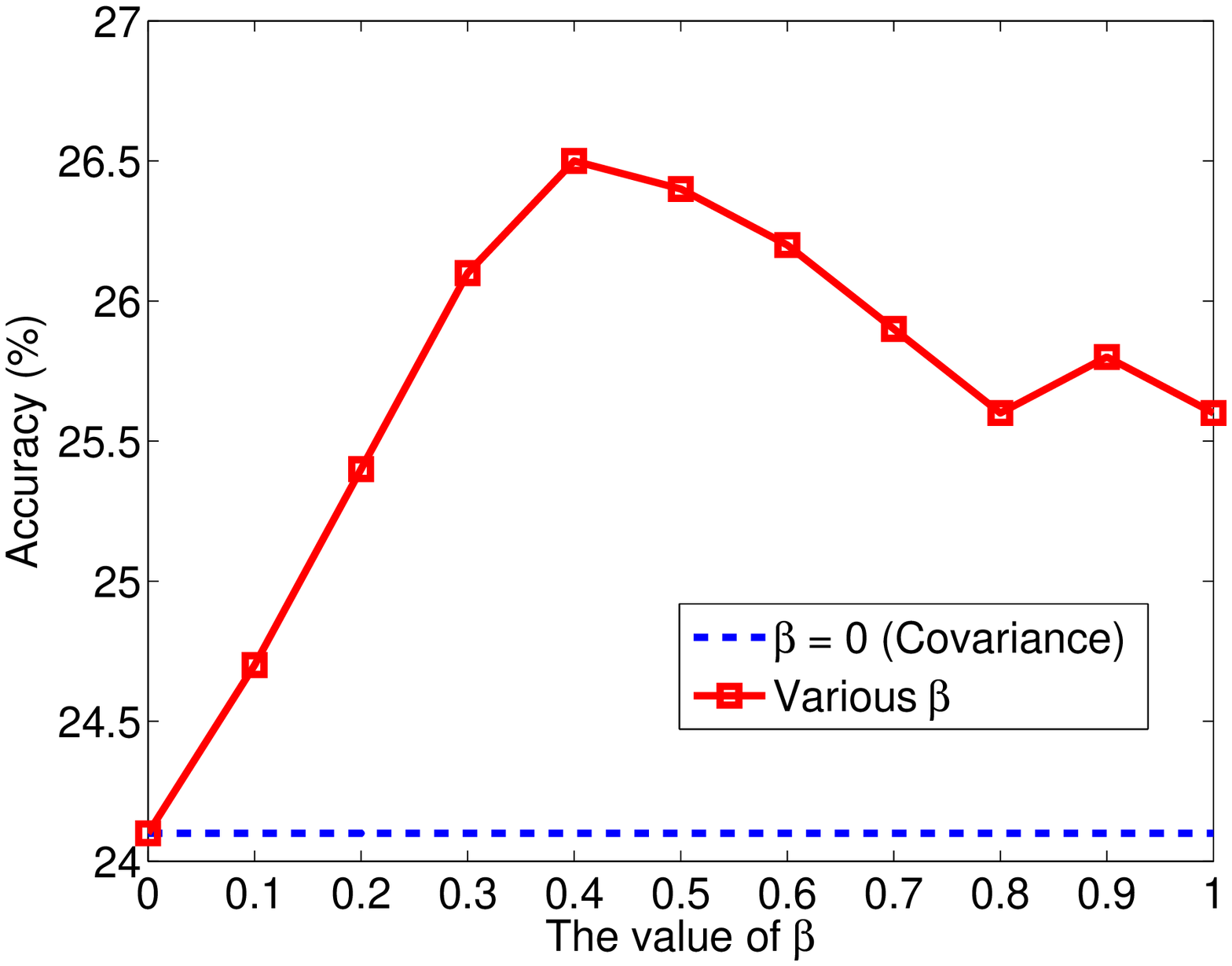}
\end{minipage}}%
\subfigure{\label{figure:LRSVM}
\begin{minipage}[b]{0.5\linewidth}
\centering
\includegraphics[width=1.0\textwidth]{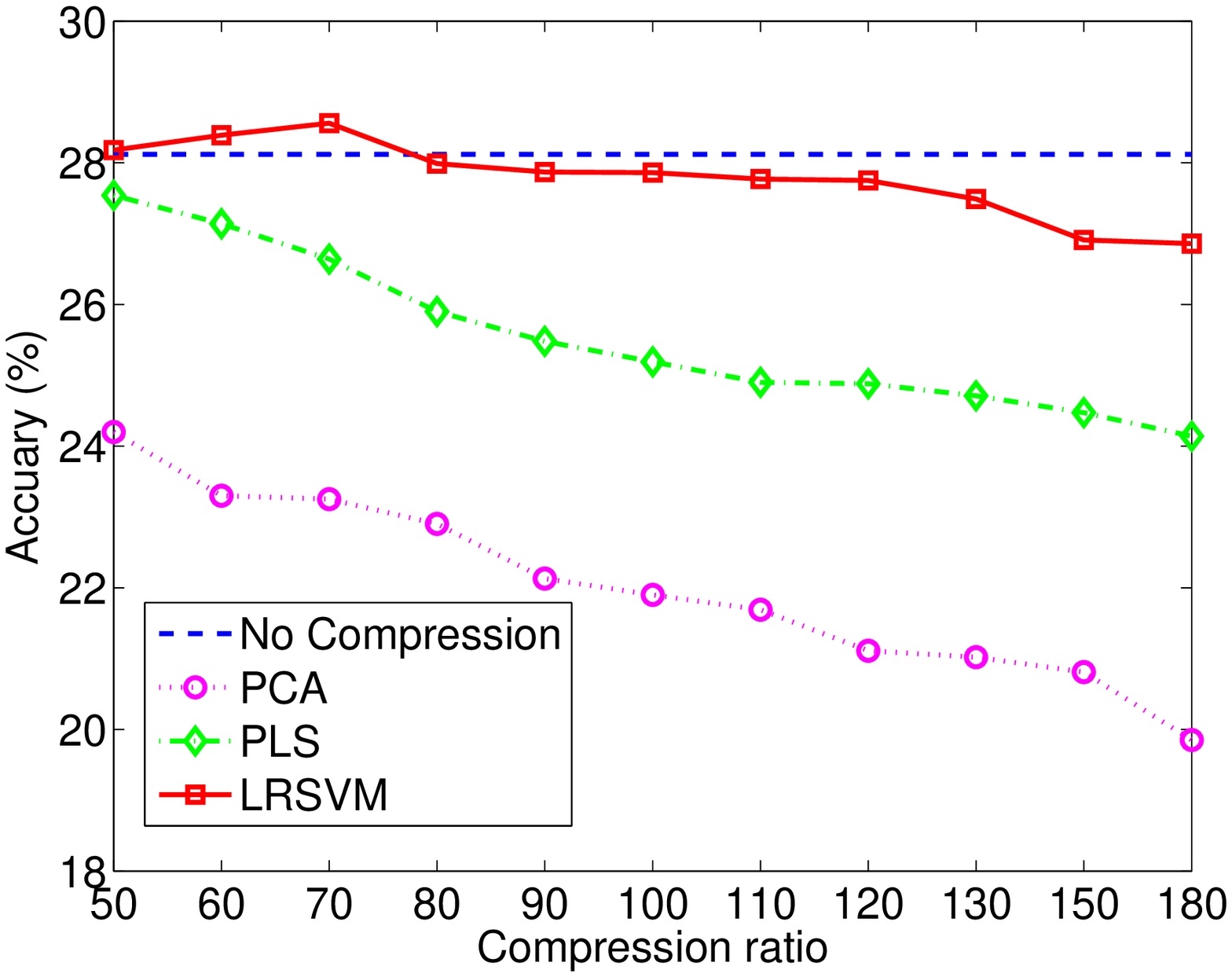}
\end{minipage}}%
\caption{Effect of balance parameter $\beta$ in Eq. (\ref{equ:embeding4}) (left) and comparison of PCA, PLS and our
LRSVM with various compression ratios on CUB200-2011 (right).}
\label{figure:exper2} 
\end{figure}

\noindent \textbf{Two well-motivated parameters}\; The proposed EPN (\ref{EPN}) is a generic method for robust estimation of covariance in high dimension space. We set parameter $\rho$ in EPN (\ref{EPN}) as $0.5$ in all databases. From Table \ref{table:parameters}, we can see that EPN can bring $1.2\%$ performance gain over the relevant method without EPN. The embedding parameter $\beta$ (\ref{equ:embeding5}) balances the effect of mean vector and covariance matrix. To test its effect, we determine the optimal value of $\beta$ via cross validation. The performances of our CLM with various $\beta$ are illustrated in Fig. \ref{figure:exper2} (left). Compared to $\beta=0$ (covariance matrix only \cite{carreira_eccv12,carreira_pami14}) and $\beta=1$ (the embedding in \cite{RePEcjmvana}), appropriate balancing at $\beta=0.4$ achieves $2.4\%$ and $0.9\%$ gains, respectively.

\noindent \textbf{LRSVM}\; To evaluate the proposed LRSVM method, we compare LRSVM with unsupervised principal component analysis (PCA) and supervised partial least square (PLS) \cite{ArenasGarciaPH06} under different compression ratios. The LRSVM is initialized by PCA, and the results on CUB200-2011 are illustrated in Fig. \ref{figure:exper2} (right). From it we can see that LRSVM always performs better than PLS, and is superior to PCA by a large margin. Different from PLS which exploits the least squares loss, LRSVM uses the hinge loss. We argue that the improvement owes to the joint learning of dimensionality reduction and classifier. Note that, with larger compression ratio, LRSVM achieves larger improvement over PCA and PLS. Meanwhile, the proposed LRSVM has insignificant performance loss (less than $1.5\%$) with large compression ratio ($> 100$). We also can see that LRSVM can slightly improve the performance of our CLM when compression rations are smaller ($<80$), which we  owe to that LRSVM can suppress some noisy information. In general, we set compression ratio as $80 \sim 100$ to balance the efficiency and effectiveness.

\noindent \textbf{Impact of PBR}\; We apply PBR to CUB200-2011 and the results are presented in Table \ref{table:parameters}. We can see that the method using PBR achieves great gains (more than $7.5\%$) over the one without PBR. Note that we achieve about $1\%$ gain in VOC2007 by using PBR. It shows that our PBR is a general method to handle background for CLM. The gains achieved by using ground truth (GT) bounding box indicate more advanced background removal methods have further ability to improve the recognition performance of our CLM. Compared with the improvement in CUB200-2011, the gains in VOC2007 are relative small. The reasons are mainly that the saliency-based methods fail to locate precisely the foregrounds in the challenging databases, and CUB200-2011 only contains one object per image while one image may contain multiple objects in VOC2007. PBR can not segment image into multiple objects so that multi-object images will heavily influence the performance of CLM.

\begin{table*}
\begin{center}
\footnotesize
\scriptsize
\begin{tabular}{|l|c|c|c|c|c|c|c|c|c|c|}
\hline
Database & Classes & Images in total & Training/Test & Measurement & Scale & View & Illumination & Pose & Bg Clutter& Occlusion\\
\hline
CUB200-2011 \cite{WahCUB2002011} & 200 & 11,788 & Split in \cite{WahCUB2002011}& Acc. of split & $\checkmark$ & $\checkmark$ & $\checkmark$& $\checkmark$ & $\checkmark$ & $\checkmark$\\
\hline
Caltech101 \cite{TPAMI2006} & 102 & 9,144 & 30/remaining per class & Acc. of 5 runs & $\checkmark$ & & & $\checkmark$ & &\\
\hline
Caltech256 \cite{Caltech256} & 256 & 30,607 & 30/remaining per class & Acc. of 5 runs & $\checkmark$ & & & $\checkmark$ & & $\checkmark$\\
\hline
Sports8 \cite{LiF07} & 8 & 1,792 & 70/60 per class& Acc. of 5 runs &  & $\checkmark$ &$\checkmark$ &$\checkmark$ & $\checkmark$ & \\
\hline
KTH-TIPS2b \cite{CaputoHM05} & 11 & 4,752 & \cite{Cimpoi_2014_CVPR} & Acc. of splits & $\checkmark$ & $\checkmark$ &$\checkmark$ & & &\\
\hline
FMD \cite{FMD} & 10 & 1,000 & 50/50 per class& Acc. of 5 runs & $\checkmark$ & $\checkmark$ &$\checkmark$ & & & \\
\hline
VOC2007 \cite{Everingham10} & 20 & 9,963 & Split in \cite{Everingham10} & mAP of split & $\checkmark$ & $\checkmark$ & $\checkmark$& $\checkmark$ & $\checkmark$ & $\checkmark$\\
\hline
Scene15 \cite{Lazebnik:2006:BBF} & 15 & 4,485 & 100/remaining per class & Acc. of 5 runs&  & $\checkmark$ & $\checkmark$& & $\checkmark$ &\\
\hline
\end{tabular}
\end{center}
\caption{Descriptions and experimental setup on eight widely used benchmarks.}
\label{table:database}
\end{table*}

\begin{table*}[t]
\scriptsize
\centering
\subtable[CUB200-2011]{
\begin{minipage}[t]{0.22\textwidth}
\label{tab:CUB200}
\begin{tabular}{l|c}
\hline
Methods & Acc.  \\
\hline
BoF-hard \cite{Lazebnik:2006:BBF} & $18.6$ \\
FV \cite{sanchez} & $25.8$ \\
FV + eSIFT & $27.3$ \\
Kobayashi2014 \cite{Kobayashi_2014_CVPR} & $27.3$ \\
PPK \cite{ZhangFD12} & $28.2$ \\
\hline
CLM (SIFT)& $18.6$ \\
CLM (eSIFT) & $28.1$\\
CLM (LogCov) & $19.1$\\
CLM (eLogCov) & $28.6$\\
CLM (eSIFT) + PBR & $36.0$\\
\hline
\end{tabular}
\end{minipage}}
\subtable[Caltech101]{
\begin{minipage}[t]{0.25\textwidth}
\label{tab:Caltech101}
\begin{tabular}{l|c}
\hline
Methods & Acc. (Tr. = 30) \\
\hline
FV+SIFT \cite{sanchez} & $80.8\pm0.3$ \\
FV+eSIFT  & $83.7\pm0.3$ \\
DeCAF \cite{DonahueJVHZTD14} & $86.9\pm0.7$ \\
O2P+eSIFT \cite{carreira_pami14}& $80.8$ \\
SQ-O2P+SIFT \cite{NIPS2014_5286} & $79.5$ \\
NBNN \cite{Irani-Definese-CVPR2008}& $77.8\pm0.3$ \\
\hline
CLM (SIFT)& $84.9\pm0.1$ \\
CLM (eSIFT) & $86.3\pm0.3$\\
CLM (LogCov) & $82.5\pm0.3$\\
CLM (eLogCov) & $84.7\pm0.2$\\
\hline
\end{tabular}
\end{minipage}}
\subtable[Caltech256]{
\begin{minipage}[t]{0.25\textwidth}
\label{tab:Caltech256}
\begin{tabular}{l|c}
\hline
Methods & Acc. (Tr. = 30)\\
\hline
BoF-LLC \cite{WangYYLHG10} & $41.2$ \\
FV+SIFT \cite{sanchez} & $47.4\pm0.1$ \\
FV+eSIFT  & $50.1\pm0.3$ \\
Kobayashi2014 \cite{Kobayashi_2014_CVPR} & $49.8\pm0.1$ \\
NBNN \cite{Irani-Definese-CVPR2008}& $43$ \\
M-HMP \cite{Irani-Definese-CVPR2008}& $50.7$ \\
\hline
CLM (SIFT)& $48.9\pm0.2$ \\
CLM (eSIFT) & $53.6\pm0.2$\\
CLM (LogCov) & $48.6\pm0.3$\\
CLM (eLogCov) & $53.2\pm0.1$\\
\hline
\end{tabular}
\end{minipage}}
\subtable[Sports8]{
\begin{minipage}[t]{0.25\textwidth}
\label{tab:Scene15}
\begin{tabular}{l|c}
\hline
Methods & Acc. \\
\hline
FV+SIFT \cite{sanchez} & $91.3\pm1.3$ \\
FV+eSIFT  & $90.4\pm1.2$ \\
Kobayashi2014 \cite{Kobayashi_2014_CVPR} & $92.6\pm0.7$ \\
GG (ad-linear) \cite{Nakayama-CVPR2010} & $80.2$ \\
GG (ct-linear) \cite{Nakayama-CVPR2010} & $82.9\pm1.0$ \\
GG + KL Div. \cite{Nakayama-CVPR2010} & $84.4\pm1.4$ \\
\hline
CLM (SIFT)& $88.8\pm1.0$ \\
CLM (eSIFT) & $91.5\pm1.2$\\
CLM (LogCov) & $88.3\pm1.3$\\
CLM (eLogCov) & $90.7\pm0.7$\\
\hline
\end{tabular}
\end{minipage}}

\subtable[KTH-TIPS2b]{
\begin{minipage}[t]{0.24\textwidth}
\label{tab:KTH-TIPS2b}
\begin{tabular}{l|c}
\hline
Methods & Acc. \\
\hline
BoF-LLC \cite{WangYYLHG10} & $57.6\pm2.3$ \\
VLAD \cite{JDSP10} & $63.1\pm1.0$ \\
FV+SIFT \cite{sanchez} & $69.3\pm1.0$ \\
FV+eSIFT  & $71.3\pm3.1$ \\
DeCAF \cite{DonahueJVHZTD14} & $70.7\pm1.7$ \\
Attributes \cite{Cimpoi_2014_CVPR} & $73.8\pm 1.3 $ \\
\hline
CLM (SIFT)& $71.8\pm3.1$ \\
CLM (eSIFT) & $75.2\pm2.6$\\
CLM (LogCov) & $72.2\pm3.3$\\
CLM (eLogCov) & $73.6\pm2.6$\\
\hline
\end{tabular}
\end{minipage}}
\subtable[FMD]{
\begin{minipage}[t]{0.25\textwidth}
\label{tab:FMD}
\begin{tabular}{l|c}
\hline
Methods & Acc. \\
\hline
VLAD \cite{JDSP10} & $52.6\pm1.5$ \\
FV+SIFT \cite{sanchez} & $58.3\pm1.0$ \\
FV+eSIFT  & $58.9\pm1.7$ \\
Kobayashi2014 \cite{Kobayashi_2014_CVPR} & $57.3\pm0.9$ \\
DeCAF \cite{DonahueJVHZTD14} & $60.7\pm2.1$ \\
Attributes \cite{Cimpoi_2014_CVPR} & $61.1\pm 1.4$ \\
\hline
CLM (SIFT)& $51.6\pm1.2$ \\
CLM (eSIFT) & $57.7\pm1.6$\\
CLM (LogCov) & $62.4\pm1.5$\\
CLM (eLogCov) & $64.2\pm1.0$\\
\hline
\end{tabular}
\end{minipage}}
\subtable[VOC2007]{
\begin{minipage}[t]{0.23\textwidth}
\label{tab:VOC2007}
\begin{tabular}{l|c}
\hline
Methods & mAP. \\
\hline
BoF-LLC \cite{WangYYLHG10} & $57.4$ \\
SV \cite{ZhouYZH10} & $58.2$ \\
SQ-O2P+SIFT \cite{NIPS2014_5286} & $51.0$ \\
FV+SIFT \cite{sanchez} & $61.8$ \\
FV+eSIFT  & $60.8$ \\
Kobayashi2014 \cite{Kobayashi_2014_CVPR} & $63.8$ \\
\hline
CLM (SIFT)& $55.8$ \\
CLM (eSIFT) & $60.4$\\
CLM (LogCov) & $56.6$\\
CLM (eLogCov) & $61.7$\\
\hline
\end{tabular}
\end{minipage}}
\subtable[Scene15]{
\begin{minipage}[t]{0.24\textwidth}
\label{tab:Scene15}
\begin{tabular}{l|c}
\hline
Methods & Acc. \\
\hline
SV \cite{ZhouYZH10} & $85.0$ \\
FV+SIFT \cite{sanchez} & $88.1\pm0.2$ \\
FV+eSIFT  & $89.4\pm0.2$ \\
GG (ad-linear) \cite{Nakayama-CVPR2010} & $79.8$ \\
GG (ct-linear) \cite{Nakayama-CVPR2010} & $82.3\pm0.4$ \\
GG + KL Div. \cite{Nakayama-CVPR2010} & $86.1\pm0.5$ \\
\hline
CLM (SIFT)& $88.1\pm0.4$ \\
CLM (eSIFT) & $89.4\pm0.4$\\
CLM (LogCov) & $88.3\pm0.6$\\
CLM (eLogCov) & $89.2\pm0.5$\\
\hline
\end{tabular}
\end{minipage}}
\caption{Comparison (in \%) with state-of-the-art methods on eight widely used benchmark datasets}
\label{table:results}
\end{table*}

\subsection{Comparison with state-of-the-art methods}

We compare our CLM with more than ten state-of-the-art methods on eight widely used benchmarks. The descriptions and experimental setup on these benchmarks are listed in Table \ref{table:database}. We report the results in Table \ref{table:results}, and discuss the experimental results as follows.

\noindent \textbf{Comparison of various local descriptors}\; We combine our CLM with four kinds of local descriptors, and assess them on all databases. From Table \ref{table:results} we can see that SIFT and LogCov achieve comparable results. For object recognition, LogCov is superior to SIFT on CUB200-2011 and VOC2007 while SIFT outperforms LogCov on Caltech101 and Caltech256. On scene categorization, SIFT and LogCov obtain similar performances on both Sports8 and Sence15. For texture and material classification, SIFT achieves gains over LogCov on KTH-TIPS2b while LogCov is superior to SIFT by a large margin on FMD. The eSIFT and eLogCov perform with the similar rule as SIFT and LogCov, respectively. The enrichment on SIFT and LogCov can considerably boost the performance of our CLM, which encourages us to utilize more informative descriptors for further improvement.

\noindent \textbf{Comparison with counterparts}\; Here, we compare our CLM with its counterparts, O2P \cite{carreira_pami14}, Global Gaussian (GG) \cite{Nakayama-CVPR2010} and NBNN \cite{Irani-Definese-CVPR2008}. As shown  in Tables \ref{table:parameters} \&  \ref{table:results}, our CLM significantly outperforms O2P \cite{carreira_pami14} on CUB200-2011 and Caltech101, and is also superior to its variant with sparse quantization (SQ-O2P) \cite{NIPS2014_5286} on Caltech101 and VOC2007 by a large margin, which are mainly due to the appropriate use of mean information and EPN. Moreover, our CLM performs much better than GG methods \cite{Nakayama-CVPR2010} with ad-hoc linear kernel (ad-linear), center tangent linear kernel (ct-linear) and KL divergence on Sports8 and Sence15. The ad-linear can be seen as a baseline in Euclidean space. It is mentionable that the methods in \cite{Nakayama-CVPR2010} exploit probabilistic discriminant analysis (PDA) as a classifier. If SVM is used, their results will drop to $71.7\%$, $78.8\%$ and $81.4\%$ on Sports8, and $74.3\%$, $80.7\%$ and $83.1\%$ on Scene15, respectively. We attribute the gains of our CLM over \cite{Nakayama-CVPR2010} to the use of two-step metric with the proposed well-motivated parameters. We also compare our CLM with NBNN \cite{Irani-Definese-CVPR2008}. It is easy to see that our CLM performs much better than NBNN on Caltech101 and Caltech256. The main differences between our CLM and NBNN are that our CLM employs an effective model-to-model distance and SVM classifier.

\vspace{-0.1cm}
\noindent \textbf{Comparison with FV}\; We make a comprehensive comparison with one state-of-the-art BoF method, FV \cite{sanchez}, throughout all databases, and also adopt enrichment SIFT (eSIFT) to FV. On all databases except for FMD, our CLM achieves better than or comparable performances with FV when SIFT or eSIFT is used. On FMD, with SIFT or eSIFT, our CLM is inferior to FV, but with LogCov or eLogCov, our CLM is much better than FV. In our experiments, we find that LogCov or eLogCov is not very suitable for FV, so the relevant results are not reported. It is found that our CLM is more sensitive to local descriptors than FV, as eSIFT brings less or no gains on FV while our CLM greatly benefits from the enrichment on SIFT or LogCov.

\vspace{-0.1cm}
\noindent \textbf{Comparison with other state-of-the-art methods}\; Some recent results are also presented for comparison. On Caltech101, DeCAF \cite{DonahueJVHZTD14} with 6 layers CNN and dropout strategy \cite{JMLRv15} slightly outperforms our CLM. Without dropout, the result of DeCAF drops to $84.8\%$. On Caltech256, our CLM outperforms the deep architecture Multipath Hierarchical Matching Pursuit (M-HMP) \cite{bo_cvpr13}  by $2.9\%$.  Cimpoi \etal \cite{Cimpoi_2014_CVPR} achieved state-of-the-art results on KTH-TIPS2b and FMD with semantic attributes which are trained on the additional database by combining FV \cite{sanchez} and DeCAF \cite{DonahueJVHZTD14}. Our CLM is superior to the method with attributes, FV and DeCAF. By combining attribute features, FV and DeCAF, Cimpoi \etal \cite{Cimpoi_2014_CVPR} obtained $77.3\%$ and $67.1\%$ accuracy on KTH-TIPS2b and FMD. Kobayashi \cite{Kobayashi_2014_CVPR} proposed a histogram transformation method, and it achieves state-of-the-art results on Sports8 and VOC2007.

\noindent \textbf{Summary}\; In this paper, we assess our CLM on eight image benchmarks, as shown in Table
\ref{table:database}, which contains various transformations or noisy factors. We claim that (1) the results
on Caltech101 and Caltech256 show that our CLM can well deal with location and pose variations of objects; (2)
the results on FMD and KTH-TIPS2b show that our CLM is robust to scale, viewpoint, illumination and appearance
variation; (3) the results on Sports8 and Sence15 indicate our CLM can well classify scene images with
certain background clutters; and (4) the results on CUB200-2011 and VOC2007 demonstrate our CLM also can handle
images with complex surroundings, such as heavy background clutters and occlusion.
\subsection{Computational complexity analysis }

Our CLM for classification mainly consists of three components: extracting local descriptors, computing Gaussian models using Eq.(\ref{equ:embeding4}) followed by EPN (\ref{EPN}) and matrix logarithm in Eq.(\ref{distance_M_L}), and learning LRSVM for classification. Most of the computational costs of CLM lie in the eigenvalue decomposition produced by EPN and matrix logarithm. Their computational complexity are $O(k^{3})$ and $O((k+1)^{3})$, respectively, where $k$ is the dimension of local descriptors. During joint training of low-rank matrix and SVM classifier, optimizing the objective function (\ref{LRSVM4}) consists of alternating SVM minimization problem and trace minimization problem, whose complexity is $O(J(N^{2}D+D^{3}+Bd^{3}))$, where $N$ is the number of training samples of dimension $D=Bd$, and $J$ is the number of iterations which is less than $3$ in our experiments.

Here, we give empirical running time by taking KTH-TIPS2b and Caltech101 as examples. The time of computing image representation, which includes extraction of SIFT at multiple scales, and the time of computation of Gaussian models and embedding matrices, are 30 minutes on KTH-TIPS2b and 1.5 hours on Caltech101. The average time of modeling one image takes about 0.4 second and 0.6 second on relevant databases. For each trial, training (resp. test) of LRSVM takes 20s (resp. 2s) and 7min (resp. 40s) on KTH-TIPS2b and Caltech101, respectively.

\section{Discussion and conclusion}\label{Dis}

The bag-of-features (BoF) is a popular method in classification and recognition fields, demonstrating convincing performance in many computer vision tasks in the past years. It might seem that training codebook \&
descriptor coding are indispensable ingredients. However, the codebookless model (CLM) proposed in this work has proven to be an effective alternative method to the BoF methods for image classification. Below we give some discussions about why CLM shows such competitive performance.

Different from the BoF methods, our CLM leverages continuous functions for statistical modeling of local descriptors, which does not need codebook and thus has no quantization brought in. Recent research \cite{ChenCWS13} showed that high dimensionality can bring impressive performance. The state-of-the-art BoF methods such as SV/VLAD or FV have inherently high dimensionality, which, in our opinion, is the key for characterizing distinctness and discriminativess of individual images as well as image categories. Our CLM directly employs the first- and second-order statistics of high dimensional local descriptors, giving rise to informative image-level models of high dimensionality as well. In this respect, it is worthwhile to study more informative or high dimensional CLM. Moreover, as shown in \cite{carreira_eccv12,carreira_pami14}, the CLM is more efficient than the BoF methods for modeling images because learning codebook \& coding are not necessary.  In addition, the CLM may be more suitable for the tasks where the datasets will be regularly updated or increased, and thus the codebook in the BoF model has to be regularly adjusted to fit the changing datasets.

The contributions of this paper are concluded as follows. (1) Our work has clearly shown that the CLM is a very competitive alternative to the mainstream BoF model. We hope our work can raise potential interests in the classification (or retrieval) community and pave a way to future research. (2) Our method enables Gaussian models to be successfully combined with linear SVM classifier, which makes our method scalable and efficient. The key is that we embed Gaussian models into a vector space which also allows us to perform joint low-rank learning and SVM on Gaussian manifold. Meanwhile, the proposed two well-motivated parameters further improve our CLM. (3) We performed extensive experiments, evaluating various aspects of our CLM and comparing with its counterparts as well as state-of-the-art methods. The comprehensive experiments demonstrated the promising performance of our CLM.

{\small
\bibliographystyle{ieee}
\bibliography{egbib}

\begin{thebibliography}{10}\itemsep=-1pt

\bibitem{ArenasGarciaPH06}
J.~Arenas-Garc¨ªa, K.~B. Petersen, and L.~K. Hansen.
\newblock Sparse kernel orthonormalized {PLS} for feature extraction in large
  data sets.
\newblock In {\em NIPS}, 2006.

\bibitem{Arsigny:2005}
V.~Arsigny, P.~Fillard, X.~Pennec, and N.~Ayache.
\newblock Fast and simple calculus on tensors in the {L}og-{E}uclidean
  framework.
\newblock In {\em MICCAI}, 2005.

\bibitem{BIK11}
C.~Beecks, A.~M. Zimmer, S.~Kirchhoff, and T.~Seidl.
\newblock Modeling image similarity by gaussian mixture models and the
  signature quadratic form distance.
\newblock In {\em ICCV}, 2011.

\bibitem{bo_cvpr13}
L.~Bo, X.~Ren, and D.~Fox.
\newblock Multipath sparse coding using hierarchical matching pursuit.
\newblock In {\em CVPR}, 2013.

\bibitem{NIPS2009_3874}
L.~Bo and C.~Sminchisescu.
\newblock Efficient match kernel between sets of features for visual
  recognition.
\newblock In {\em NIPS}, 2009.

\bibitem{Irani-Definese-CVPR2008}
O.~Boiman, E.~Shechtman, and M.~Irani.
\newblock In defense of nearest-neighbor based image classification.
\newblock In {\em CVPR}, 2008.

\bibitem{NIPS2014_5286}
X.~Boix, G.~Roig, S.~Diether, and L.~V. Gool.
\newblock Self-adaptable templates for feature coding.
\newblock In {\em NIPS}, 2014.

\bibitem{CaputoHM05}
B.~Caputo, E.~Hayman, and P.~Mallikarjuna.
\newblock Class-specific material categorisation.
\newblock In {\em ICCV}, 2005.

\bibitem{carreira_eccv12}
J.~Carreira, R.~Caseiro, J.~Batista, and C.~Sminchisescu.
\newblock {Semantic Segmentation with Second-Order Pooling}.
\newblock In {\em ECCV}, 2012.

\bibitem{carreira_pami14}
J.~Carreira, R.~Caseiro, J.~Batista, and C.~Sminchisescu.
\newblock {Free-Form Region Description with Second-Order Pooling}.
\newblock {\em TPAMI}, PP:1, 2014.

\bibitem{Chang:2011}
C.-C. Chang and C.-J. Lin.
\newblock {LIBSVM}: A library for support vector machines.
\newblock {\em ACM TIST}, 2(3):27, 2011.

\bibitem{ChenCWS13}
D.~Chen, X.~Cao, F.~Wen, and J.~Sun.
\newblock Blessing of dimensionality: High-dimensional feature and its
  efficient compression for face verification.
\newblock In {\em CVPR}, 2013.

\bibitem{Cimpoi_2014_CVPR}
M.~Cimpoi, S.~Maji, I.~Kokkinos, S.~Mohamed, and A.~Vedaldi.
\newblock Describing textures in the wild.
\newblock In {\em CVPR}, 2014.

\bibitem{DonahueJVHZTD14}
J.~Donahue, Y.~Jia, O.~Vinyals, J.~Hoffman, N.~Zhang, E.~Tzeng, and T.~Darrell.
\newblock Decaf: {A} deep convolutional activation feature for generic visual
  recognition.
\newblock In {\em ICML}, 2014.

\bibitem{Donoho2014}
D.~L. Donoho, M.~Gavish, and I.~M. Johnstone.
\newblock Optimal shrinkage of eigenvalues in the spiked covariance model.
\newblock {\em arXiv}, 1311.0851, 2014.

\bibitem{Dryden2009}
L.~Dryden, A.~Koloydenko, and D.~Zhou.
\newblock Non-euclidean statistics for covariance matrices, with applications
  to diffusion tensor imaging.
\newblock {\em Annals of Applied Statistics}, 2009.

\bibitem{Everingham10}
M.~Everingham, L.~Van~Gool, C.~K.~I. Williams, J.~Winn, and A.~Zisserman.
\newblock The {P}ascal {V}isual {O}bject {C}lasses {(VOC)} {C}hallenge.
\newblock {\em IJCV}, 88(2):303--338, 2010.

\bibitem{TPAMI2006}
L.~Fei-Fei, R.~Fergus, and P.~Perona.
\newblock One-shot learning of object categories.
\newblock {\em TPAMI}, 28(4):594--611, 2006.

\bibitem{GongWang2009}
L.~Gong, T.~Wang, and F.~Liu.
\newblock Shape of gaussians as feature descriptors.
\newblock In {\em CVPR}, 2009.

\bibitem{Grauman05thepyramid}
K.~Grauman and T.~Darrell.
\newblock The pyramid match kernel: Discriminative classification with sets of
  image features.
\newblock In {\em ICCV}, 2005.

\bibitem{Caltech256}
G.~Griffin, A.~Holub, and P.~Perona.
\newblock {The Caltech-256}.
\newblock Technical report, California Institute of Technology, 2007.

\bibitem{FMD}
L.~haran, R.~Rosenholtz, and E.~H. Adelson.
\newblock Material perception: What can you see in a brief glance?
\newblock {\em Jour. of Vis.}, 9(8):784, 2009.

\bibitem{Mehrtash14}
M.~T. Harandi, M.~Salzmann, and R.~Hartley.
\newblock From manifold to manifold: Geometry-aware dimensionality reduction
  for spd matrices.
\newblock In {\em ECCV}, 2014.

\bibitem{Jayasumana_2013}
S.~Jayasumana, R.~Hartley, M.~Salzmann, H.~Li, and M.~Harandi.
\newblock Kernel methods on the riemannian manifold of symmetric positive
  definite matrices.
\newblock In {\em CVPR}, 2013.

\bibitem{JDSP10}
H.~J\'egou, M.~Douze, C.~Schmid, and P.~P\'erez.
\newblock Aggregating local descriptors into a compact image representation.
\newblock In {\em CVPR}, 2010.

\bibitem{JiY09}
S.~Ji and J.~Ye.
\newblock Linear dimensionality reduction for multi-label classification.
\newblock In {\em IJCAI}, 2009.

\bibitem{Jiang_2013_ICCV}
B.~Jiang, L.~Zhang, H.~Lu, C.~Yang, and M.-H. Yang.
\newblock Saliency detection via absorbing markov chain.
\newblock In {\em ICCV}, 2013.

\bibitem{Kobayashi_2014_CVPR}
T.~Kobayashi.
\newblock Dirichlet-based histogram feature transform for image classification.
\newblock In {\em CVPR}, 2014.

\bibitem{koniusz:hal-00922524}
P.~Koniusz, F.~Yan, P.-H. Gosselin, and K.~Mikolajczyk.
\newblock {Higher-order Occurrence Pooling on Mid- and Low-level Features:
  Visual Concept Detection}.
\newblock Technical report, 2013.

\bibitem{Lazebnik:2006:BBF}
S.~Lazebnik, C.~Schmid, and J.~Ponce.
\newblock Beyond bags of features: Spatial pyramid matching for recognizing
  natural scene categories.
\newblock In {\em CVPR}, 2006.

\bibitem{LiF07}
L.-J. Li and F.-F. Li.
\newblock What, where and who? classifying events by scene and object
  recognition.
\newblock In {\em ICCV}, 2007.

\bibitem{LiWZ13}
P.~Li, Q.~Wang, and L.~Zhang.
\newblock A novel earth mover's distance methodology for image matching with
  gaussian mixture models.
\newblock In {\em ICCV}, 2013.

\bibitem{RePEcjmvana}
M.~Lovric, M.~Min-Oo, and E.~A. Ruh.
\newblock Multivariate normal distributions parametrized as a riemannian
  symmetric space.
\newblock {\em JMVA}, 74(1):36--48, 2000.

\bibitem{Lowe04}
D.~G. Lowe.
\newblock Distinctive image features from scale-invariant keypoints.
\newblock {\em IJCV}, 60(2):91--110, 2004.

\bibitem{Nakayama-CVPR2010}
H.~Nakayama, T.~Harada, and Y.~Kuniyoshi.
\newblock Global gaussian approach for scene categorization using information
  geometry.
\newblock In {\em CVPR}, 2010.

\bibitem{LogMetricsIJCV06}
X.~Pennec, P.~Fillard, and N.~Ayache.
\newblock A riemannian framework for tensor computing.
\newblock {\em IJCV}, pages 41--66, 2006.

\bibitem{Pratt2007}
W.~K. Pratt.
\newblock {\em \it Digital Image Processing, 4th Edition}.
\newblock John Wiley \& Sons, Inc., New York, NY, USA, 2007.

\bibitem{rubner}
Y.~Rubner, C.~Tomasi, and L.~J. Guibas.
\newblock The {E}arth {M}over's {D}istance as a metric for image retrieval.
\newblock {\em IJCV}, 40(2):99--121, 2000.

\bibitem{sanchez}
J.~Sanchez, F.~Perronnin, T.~Mensink, and J.~Verbeek.
\newblock Image classification with the {F}isher vector: Theory and practice.
\newblock {\em IJCV}, 105(3):222--245, 2013.

\bibitem{Sivic03}
J.~Sivic and A.~Zisserman.
\newblock {Video Google}: {A} text retrieval approach to object matching in
  videos.
\newblock In {\em ICCV}, 2003.

\bibitem{JMLRv15}
N.~Srivastava, G.~Hinton, A.~Krizhevsky, I.~Sutskever, and R.~Salakhutdinov.
\newblock Dropout: A simple way to prevent neural networks from overfitting.
\newblock {\em JMLR}, 15:1929--1958, 2014.

\bibitem{Stein1986}
C.~Stein.
\newblock Lectures on the theory of estimation of many parameters.
\newblock {\em Jour. of Math. Sci.}, 34(1):1373--1403, 1986.

\bibitem{Sydorov_2014_CVPR}
V.~Sydorov, M.~Sakurada, and C.~H. Lampert.
\newblock Deep fisher kernels - end to end learning of the {F}isher kernel
  {GMM} parameters.
\newblock In {\em CVPR}, 2014.

\bibitem{tuzel:region}
O.~Tuzel, F.~Porikli, and P.~Meer.
\newblock Region covariance: A fast descriptor for detection and
  classification.
\newblock In {\em ECCV}, 2006.

\bibitem{GemertVSG10}
J.~van Gemert, C.~J. Veenman, A.~W.~M. Smeulders, and J.-M. Geusebroek.
\newblock Visual word ambiguity.
\newblock {\em TPAMI}, 32(7):1271--1283, 2010.

\bibitem{vedaldi08vlfeat}
A.~Vedaldi and B.~Fulkerson.
\newblock {VLFeat}: An open and portable library of computer vision algorithms.
\newblock \url{http://www.vlfeat.org/}, 2008.

\bibitem{WahCUB2002011}
C.~Wah, S.~Branson, P.~Welinder, P.~Perona, and S.~Belongie.
\newblock {The Caltech-UCSD Birds-200-2011 Dataset}.
\newblock Technical report, 2011.

\bibitem{WangYYLHG10}
J.~Wang, J.~Yang, K.~Yu, F.~Lv, T.~S. Huang, and Y.~Gong.
\newblock Locality-constrained linear coding for image classification.
\newblock In {\em CVPR}, 2010.

\bibitem{Weston2011}
J.~Weston, S.~Bengio, and N.~Usunier.
\newblock Wsabie: Scaling up to large vocabulary image annotation.
\newblock In {\em IJCAI}, 2011.

\bibitem{YaoEtalCVPR12}
B.~Yao, G.~Bradski, and L.~Fei-Fei.
\newblock A codebook-free and annotation-free approach for fine-grained image
  categorization.
\newblock In {\em CVPR}, 2012.

\bibitem{ZhangFD12}
N.~Zhang, R.~Farrell, and T.~Darrell.
\newblock Pose pooling kernels for sub-category recognition.
\newblock In {\em CVPR}, 2012.

\bibitem{ZhouYLWLT14}
W.~Zhou, M.~Yang, H.~Li, X.~Wang, Y.~Lin, and Q.~Tian.
\newblock Towards codebook-free: Scalable cascaded hashing for mobile image
  search.
\newblock {\em TMM}, 16(3):601--611, 2014.

\bibitem{ZhouYZH10}
X.~Zhou, K.~Yu, T.~Zhang, and T.~S. Huang.
\newblock Image classification using super-vector coding of local image
  descriptors.
\newblock In {\em ECCV}, 2010.

\end{thebibliography}
}

\end{document}